\documentstyle[colap,gb4e]{article}

\newcommand{\into}{$\rightarrow$}

\title{\vspace*{-15mm}Classifiers in Japanese-to-English Machine Translation}

\author{Francis Bond \and Kentaro  Ogura \and Satoru Ikehara\footnotemark[2]
  \\ {\bf NTT Communication Science Laboratories} \\ 1-2356 Take,
  Yokosuka-shi, Kanagawa-ken, {\sc Japan} 238-03 \\ 
  {\tt  bond@nttkb.ntt.jp}\\[2mm]
  COLING~'96, August 1996\footnotemark[1]}

\begin{document}
\maketitle 

\begin{abstract}
  This paper proposes an analysis of classifiers into four major
  types: {\sc unit}, {\sc metric}, {\sc group} and {\sc species},
  based on properties of both Japanese and English.  The analysis
  makes possible a uniform and straightforward treatment of noun
  phrases headed by classifiers in Japanese-to-English machine
  translation, and has been implemented in the MT system {\bf
    ALT-J/E}.  Although the analysis is based on the characteristics
  of, and differences between, Japanese and English, it is shown to be
  also applicable to the unrelated language Thai.
\end{abstract}

\renewcommand{\thefootnote}{\fnsymbol{footnote}}
\footnotetext[1]{This paper was presented at COLING~'96 and
  appears in the proceedings: Vol I, pp 125--130.}
\footnotetext[2]{Now at Tottori University.}
\renewcommand{\thefootnote}{\arabic{footnote}}

\section{Introduction}

Noun phrases in Japanese differ from those in English in two important
ways.  First, Japanese has no equivalent syntactic category to English
determiners. Second, there is no grammatical marking of
number.\footnote{Japanese does not have contrasting singular and
  plural forms of nouns.} Because of these differences, numerical
expressions are realized very differently in Japanese and English.  In
English, countable nouns can be directly modified by a numeral: {\it 2
  dogs}.  In Japanese, however, numerals cannot directly modify common
nouns, instead a classifier is used, in the same way that a partitive
noun is used with an uncountable noun in English: {\it 2 pieces of
  furniture}.  In addition, when Japanese is translated into English,
the selection of appropriate determiners, such as articles and
possessive pronouns, and the determination of countability and number
is problematic.  

Various solutions to the problems of generating articles and
possessive pronouns and determining countability and number have been
proposed \cite{Murata:1993a,Cornish:1994,Bond:1995b}.  The differences
between the way numerical expressions are realized in Japanese and
English has been less studied \cite{Asahioka:1990}.  In this paper we
propose an analysis of classifiers based on properties of both
Japanese and English.  Our category of classifier includes both
Japanese {\it jos\=ushi\/} `numeral classifiers' and English partitive
nouns.  We divide classifiers into four major types: {\sc unit}, {\sc
  metric}, {\sc group} and {\sc species}.  {\sc unit} classifiers are
further divided into {\sc general}, {\sc typical} and {\sc special},
while {\sc metric} classifiers are divided into {\sc measure} and {\sc
  container} classifiers.  Although our analysis was based on the
characteristics of, and differences between, Japanese and English, we
found it to be strikingly similar to the analysis for Thai proposed by
Sornlertlamvanich et al. (1994), which suggests that the results may
be useful for examining other languages.

The analysis introduced in this paper has been implemented in NTT
Communication Science Laboratories' Japanese-to-English machine
translation system {\bf ALT-J/E} \cite{Ikehara:1991,Ogura:1993} since
1994.  Examples of how it has been implemented in {\bf ALT-J/E} are
woven throughout the text, although the analysis itself is not tied to
any formalism or particular representation, so is adaptable to any
system.

We start off by examining monolingual analyses of Japanese classifiers
and English partitive expressions (Section~\ref{sec:mono}).  Then we
introduce our bilingual analysis of classifiers and show how this
analysis can be used in a Japanese-to-English machine translation
system (Section~\ref{sec:comb}).  We also examine more complex cases
where classifiers are used like normal nouns (Section~\ref{sec:when}).
Finally we compare our analysis to other people's
(Section~\ref{sec:other}).

Throughout the paper we use the following abbreviations: A, B or N:
noun or noun phrase; C: classifier, X: Numeral, with Japanese in
italics.


\section{Monolingual Analyses of Classifiers}
\label{sec:mono}

\subsection{Japanese `Classifiers'}
\label{sec:jpn}

Japanese is a numeral classifier language \cite{Allan:1977}, in which
classifiers are obligatory in many expressions of quantity.  We will
refer to prototypical Japanese classifiers as {\it jos\=ushi\/}
`numerical classifiers'.

Syntactically, {\it jos\=ushi\/} are a subclass of nouns
\cite{Miyazaki:1995}. The main property distinguishing them from
normal nouns is that they can postfix to numerals, the quantifier {\it
  s\=u\/} `some' or the interrogative {\it nani\/} `what', to form a
noun phrase.  Unlike normal nouns in Japanese, {\it jos\=ushi\/} can
not form grammatical noun phrases on their own.\footnote{There are
  some examples of words that can be either a common noun or {\it
    jos\=ushi\/}: for example {\it gy\=o\/} `line' or {\it hako\/}
  `box', which can follow a numeral or stand alone. These nouns can be
  handled in two ways: (a) as a lexical class that combines the
  properties of common nouns and {\it jos\=ushi}, or (b) as two
  separate lexical entities.  {\bf ALT-J/E} follows option (b), such
  nouns are entered into the lexicon twice, once as a common noun and
  once as a {\it jos\=ushi}.}

\begin{exe}
  \ex {\it 2-hiki\/} `2 animals' \hfill (Numeral)
  \ex {\it s\=u-hiki\/} `some animals' \hfill (Quant.)
  \ex {\it nan-hiki\/} `how many animals' \hfill (Int.)
\end{exe}

The resulting numeral-classifier noun phrase can modify another noun
phrase, either linked by {\it no\/} `of' {\it `XC-no-N'}, or
`floating' elsewhere in the sentence, typically directly after the
noun phrase it modifies `{\it NXC'}.  It can also occur on its own, with
anaphoric or deictic reference.  \newcite{Asahioka:1990} identify
seven different patterns of use.  In order to concentrate on
the translation of classifiers and number, we will restrict our discussion to
noun phrases of the type {\it `XC-no-N'\/} and not discuss the
problems of resolving anaphoric reference and floating quantifiers.


Semantically, each classifier relates to a class of nouns
\cite[25]{Kuno:1973}, often fairly arbitrarily.  For example {\it -hiki\/}
`(small) animal' is used to count small animals excluding rabbits,
which are counted with {\it -wa\/} `bird'.  There is a default
classifier {\it -tsu\/} `piece' which can be used to count almost
anything.

\subsection{English `Classifiers'}
\label{sec:eng}

In English, numerals can directly modify countable nouns `X N'.  In
order to enumerate uncountable nouns, either the uncountable nouns
have to be reclassified as countable nouns, or embedded in a partitive
construction: {\it two beers\/} or {\it two cans of beer\/} `X N' or `X C
{\it of\/} N' \cite[249]{Quirk:1985}.  This partitive construction is
similar to the Japanese quantifying construction {\it `XC-no-N'}.

\newcite[249--51]{Quirk:1985} divide partitive nouns
into three main categories {\sc quality partitives}, {\sc quantity
  partitives}, and {\sc measure partitives}.  {\sc quantity
  partitives} are further divided into three cases, the first where the
embedded noun phrase is uncountable, the second where it is plural, and
the third where it is singular and countable.  All the partitive nouns
themselves are fully countable.

{\sc quantity partitives} where the embedded noun phrase is headed by
an uncountable noun, the first case, are then divided into {\sc general
  partitives} such as {\it piece\/} which serve only to quantify and
{\sc typical partitives} such as {\it grain\/} which are more
descriptive.

\section{A Bilingual Analysis of Classifiers}
\label{sec:comb}

As there is no direct fit between English and Japanese, it is
necessary to categorize the Japanese and English classifiers and to
define rules which will enable effective machine translation.  We
divide classifiers into four major types: {\sc unit}
(Section~\ref{sec:unit}), {\sc metric} (Section~\ref{sec:metric}),
{\sc group} (Section~\ref{sec:group}) and {\sc species}
(Section~\ref{sec:species}).  The main criteria for the analysis are
the restrictions placed, in English, on the countability and number of
the embedded noun phrase in a partitive construction.  Whether a noun
is a classifier, and if so which type, is marked in the lexicon for
each Japanese/English noun pair.

We distinguish between five major different noun countability
preferences, based on the analysis of \newcite{Allan:1980}, adapted
for use in machine translation by \newcite{Bond:1994}.
`Fully countable' nouns, such as {\it knife}, have both singular and
plural forms, and cannot be used with determiners such as {\it much}.
`Uncountable' nouns, such as {\it furniture}, have no plural form, and
can be used with {\it much}.  Between these two extremes are nouns
such as {\it cake}, which can be used in both countable and
uncountable noun phrases.  They have both singular and plural forms,
and can also be used with {\it much}. We divide such nouns into two
groups: `strongly countable', those that are more often used to refer
to discrete entities, such as {\it cake}, and `weakly countable',
those that are more often used to refer to unbounded referents, such
as {\it beer}.  The fifth major type of countability preference is
`pluralia tantum': nouns that only have a plural form, such as {\it
  scissors}.

\subsection{Unit classifiers}
\label{sec:unit}                               

\begin{table*}[tbp]
  \caption{Unit Classifiers} \label{tab:unit}
  \center
  \begin{tabular}{|l|l|l|l|}
    \hline
    \bf Noun Type & \bf General & \bf Typical  & \bf Special \\
    \hline
    Fully Countable & 1 dog                         &
    1 dog           & 1 slice of dog \\
    Strongly Countable  & 1 cake                        &
    1 crumb of cake    & 1 slice of cake \\
    Weakly Countable & 1 hair                        &
    1 strand of hair    & 1 slice of hair \\
    Uncountable     & 1 piece of information        &
    1 grain of information & 1 slice of information \\
    Pluralia Tantum (pair) & 1 pair of scissors            &
    1 pair of scissors & --- \\
\hline
\end{tabular}
\end{table*}

{\sc unit} classifiers are the prototypical classifiers.  A {\sc unit}
classifier will be realized in Japanese as a {\it jos\=ushi}.
However, there are three possible translations of a Japanese noun
phrase of the form {\it `XC-no-N'}, where {\it C\/} is a unit classifier:

\begin{description}
\item[Individuate:] Translate as `X N', where the classifier {\it C\/}
  is not translated and the numeral directly modifies the countable
  English noun phrase: \\{\it 1-hiki-no-inu\/} `1-piece of dog' \into\ 
  {\it 1 dog}.
\item[Part:] Translate as `X C {\it of\/} N', where the classifier is
  translated by its translation equivalent (from the transfer
  dictionary) and N is uncountable (headed by a bare singular noun):\\
  {\it 1-tsubu-no-kome\/} `1-grain of rice'\\ \into\  {\it 1 grain of rice}.

\item[Default:] Translate as `X C {\it of\/} N' where the classifier
  is replaced by a default that depends on the embedded noun and N is
  uncountable.  The default is normally {\it piece}, but this can be
  over-ridden by an explicit entry for N's default classifier in the
  lexicon: \\{\it 1-tsu-no-kagu\/} `1-piece of furniture' \\
  \into {\it 1 piece of furniture}.
\end{description}

The three types of {\sc unit} classifier are summarized in
Table~\ref{tab:unit}.\footnote{If N's countability preference is
  pluralia tantum then N will never be {\bf individuat}ed.  If N is
  {\bf part}ed or {\bf default}ed there are two possibilities: either,
  if the dictionary entry for N has the default classifier {\it
    pair\/} then it will be used as the classifier or, if N has no
  default classifier, then a different translation is searched for in
  the dictionary and used instead.  If there is no non-pluralia tantum
  translation equivalent, then the translation will default to `X C
  {\it of\/} N' as above, but with N headed by a bare plural noun.}

Having established three possible translations of the `{\it
  XC-no-N'\/} construction, we can proceed to divide {\sc unit}
classifiers into three types, depending on which of the above
alternatives is most suitable.  The first, {\sc general} classifiers,
are those that have no special meaning of their own, but are used only
to quantify the denotation of a noun.  Typical examples are {\it -
  tsu\/} `piece' and {\it -ko\/} `piece'.  If N is fully, strongly or
weakly countable, then the classifier is not translated (individuate).
If N is uncountable, then the classifier is translated as the default
(default).  The second type of classifier, {\sc typical}, consists of
those classifiers which are descriptive in their own right, such as
{\it -teki\/} `drop'. If N is fully countable, then the classifier
will not be translated (individuate), otherwise the classifier is
translated (part).  The final type of classifier, {\sc special}, is
rare: classifiers which force an uncountable interpretation of even
countable nouns, for example {\it -kire\/} `slice'.  N is always {\bf
  part}ed: {\it 1-kire-no-inu\/} `1-slice of dog' \into {\it 1 slice
  of dog}.

The translation of classifiers is complicated by the fact that
classifiers and their relationships to nouns are both arbitrary and
language dependent.  Consider the Japanese classifier {\it -mai}
`sheet', which is used for counting flat objects.  This has no direct
English equivalent.  As a default, it is entered in the dictionary as
a {\sc general} classifier with the translation {\it piece}.  There
are however several flat objects for which {\it piece\/} is
inappropriate in English: food-stuffs ({\it slice\/}); paper, glass,
cloth and leather ({\it sheet\/}); bacon ({\it rasher\/}); and
financial contracts ({\it contract\/}).  The selection of an
appropriate translation is not dependent on this analysis and can be
left to the normal machine translation process.  In {\bf ALT-J/E} it
is done by examining the semantic category of the embedded noun.  Once
an appropriate translation of the classifier has been found, knowledge
of its type allows the system to decide the appropriate form of the
final translation.

\subsection{Metric classifiers}
\label{sec:metric}

\begin{table*}[htp]
\caption{Metric Classifiers}\label{tab:metric}
\center
\begin{tabular}{|l|l|l|}
\hline
\bf Noun Type & \bf Container & \bf Measure  \\
\hline
Fully Countable & 1 box of dogs                         &
1 kg of ants  \\
Strongly Countable  & 1 box of cake                 &
1 kg of cake   \\
Weakly Countable& 1 box of beer                 &
1 kg of beer    \\
Uncountable     & 1 box of furniture    &
1 kg of furniture \\
Pluralia Tantum & 1 box of scissors             &
1 kg of scissors \\
\hline
\end{tabular}

\caption{Group and Species Classifiers}\label{tab:group}
\smallskip
\begin{tabular}{|l|l||l|l|}
\hline
\bf Noun Type & \bf Group & \bf Species (Si) & \bf Species (Pl) \\
\hline
Fully Countable & 1 set of dogs               & 
1 kind of dog   & 2 kinds of dogs \\
Strongly Countable  & 1 set of cakes                & 
1 kind of cake  & 2 kinds of cakes \\
Weakly Countable& 1 set of beer               &
1 kind of beer  & 2 kinds of beer \\
Uncountable     & 1 set of information  &
1 kind of information & 2 kinds of information \\
Pluralia Tantum & 1 set of scissors           &
1 kind of scissors & 2 kinds of scissors \\
\hline
\end{tabular}

\end{table*}

The next overall category is {\sc metric} classifiers.  A noun phrase
of the form {\it `XC-no-N'}, where C is a {\sc metric} classifier
will be translated as `X C {\it of\/} N', where N will be plural if it is
headed by a fully countable or pluralia tantum noun.  We further
subdivide {\sc metric} classifiers depending on whether the resulting
English noun phrase will have singular verb agreement ({\sc measure}
classifiers), or plural verb agreement ({\sc container} classifiers)
as its default.

\begin{exe}
  \ex {\it 2-kg-no-kami-ha j\=ubun da\/} `2 kg of paper-{\sc top}
  enough is' \into\  {\it 2 kg of paper is enough\/} \label{s:meas}
  \ex {\it 2-hako-no-kami-ha j\=ubun da\/} `2 box of paper-{\sc top}
  enough is' \into\  {\it 2 boxes of paper are enough\/}\label{s:cont}
\end{exe}

In fact both (\ref{s:meas}) and (\ref{s:cont}) could be translated
with singular or plural verb agreement.  The differentiation into {\sc
  measure} and {\sc container} provides a graceful default.  Examples
are given in Table~\ref{tab:metric}.

\subsection{Group classifiers}
\label{sec:group}


{\sc group} classifiers combine with plural or uncountable noun
phrases to make a countable noun phrase representing a group or set.
A noun phrase of the form {\it `XC-no-N'}, where C is a {\sc group}
classifier will be translated as `X C {\it of\/} N', where N will be
plural if it is headed by a fully or strongly countable noun or a
pluralia tantum.  Noun phrases of the form {\it `N-no-C'}, where C is
a {\sc group} classifier (but not a {\it jos\=ushi\/}) will also be
translated as `C of N' where N will be plural if it is headed by a
fully or strongly countable noun or a pluralia tantum.  This allows us
to give a uniform treatment of noun phrases such as (\ref{s:box-1})
and (\ref{s:box-2}) during English generation, even though their
Japanese structure is very different.

\begin{exe}
  \ex \label{s:box-1} {\it 2-hako-no-pen \/} `2 box of pen'\\
  \into\  {\it 2 boxes of pens\/} \hfill {\it `XC-no-N'\/}
  \ex \label{s:box-2} {\it pen-no-hako\/} `box of pen'\\
  \into\  {\it a box of pens\/} \hfill {\it `N-no-C'\/}
\end{exe}

Whether a noun is a {\sc group} classifier or not can also be used to
help determine the number of ascriptive and appositive noun phrases.
For example, in {\bf ALT-J/E} the countability and number of two
appositive noun phrases are made to match each other, unless one
element is plural and the other is a {\sc group} classifier.  For
example, {\it many insects, a whole swarm, \ldots\/} as opposed to
{\it many insects, bees I think, \ldots\/} \cite{Bond:1995b}.
Examples of {\sc group} classifiers are given in
Table~\ref{tab:group}.

\subsection{Species classifiers}
\label{sec:species}

The last type of classifier is {\sc species} classifiers.  {\sc
  species} classifiers are partitives of quality and can occur with
countable or uncountable noun phrases.  The embedded noun phrase will
agree in number with the head noun phrase if fully or strongly
countable: {\it a kind of car, 2 kinds of cars; a kind of equipment, 2
  kinds of equipment}.  Examples of {\sc species} classifiers are
given in Table~\ref{tab:group}.

\section{When is a Classifier a Classifier?}
\label{sec:when}

\begin{table*}[htp]
    \caption{Proposed Analysis of Classifiers}
    \label{tab:overall}
  \begin{center}
 \begin{tabular}{|l|l||ll|l|l|}\hline
\multicolumn{2}{|c||}{\bf Classifier type} & \multicolumn{2}{c|}{\bf Example} & 
\bf Japanese POS & \bf English Restriction on embedded NP \\ \hline\hline
{\bf Unit} & General & {\it -tsu\/} & `piece' & {\it jos\=ushi\/} & 
 Default classifier if uncountable head, \\
& & & & & no classifier if countable  \\\cline{2-6}
& Typical & {\it -tsubu\/} & `grain' & {\it jos\=ushi\/} & 
 Translate classifier if uncountable, \\
& & & & & no classifier if countable \\ \cline{2-6}
& Special & {\it -kire\/} & `slice' & {\it jos\=ushi\/} & 
 Translate classifier, \\
&&&&&force head to be uncountable  \\\hline
{\bf Metric} & Measure & {\it -inchi\/} & `inch'  & {\it jos\=ushi\/} & 
 Plural if possible, singular agreement  \\\cline{2-6}
& Container & {\it hako\/} & `box' & noun/{\it jos\=ushi\/} & 
  Plural if possible, normal agreement   \\\hline
\multicolumn{1}{|c}{\bf Group} && {\it mure\/} & `group' & noun/{\it jos\=ushi\/} & 
  Plural if possible \\\hline
\multicolumn{1}{|c}{\bf Species} && {\it shurui\/} & `kind' & noun/{\it jos\=ushi\/} & 
 Number agrees if possible  \\\hline
\end{tabular}   

  \caption{A comparison of different analyses}
  \label{tab:comp}
  \medskip
  \begin{tabular}{|l|l||l|l|l|}\hline
\multicolumn{2}{|c||}{\bf Proposed Analysis} & 
\bf Quirk et al.  & \bf Kamei et al.& \bf Sornlertlamvanich et al.\\\hline\hline
{\bf Unit} & General & Quantity-General &  & \\ \cline{2-3}
& Typical &  \raisebox{-1.5ex}[0ex][0ex]{Quantity-Typical} & Piece   & Unit \\ \cline{2-2}
& Special &  &  & \\ \hline
{\bf Metric} & Measure & Measure & Unit & \raisebox{-1.5ex}[0ex][0ex]{Metric} \\ \cline{2-4}
& Container &  --- & Container &  \\\hline
\multicolumn{1}{|c}{\bf Group} && Quantity-Plural & Set  & Collective  \\\hline
\multicolumn{1}{|c}{\bf Species} && Quality & Kind & --- \\\hline
\multicolumn{1}{|c}{\bf (Unit) } && --- & Times & Frequency \\\hline
\multicolumn{1}{|c}{\bf (Unit) } && --- & --- & Verbal \\\hline
  \end{tabular}
  \end{center}
\end{table*}

In the analysis given above for Japanese noun phrases of the form
{\it `XC-no-N'}, we have given no consideration to the denotation of
N, except for when choosing the appropriate translation for C.  Thus
we assume that {\it `XC-no-N'\/} will be translated as `X C {\it of\/}
N' or just `X N' if N is countable, as in (\ref{class1}) or
(\ref{class2}).

\begin{exe}
\ex     {\it 1-pai-no mizu\/} `1-cup of water' \\ \into\
{\it 1 cup of  water\/} ({\sc container}) \label{class1}
\ex     {\it 1-tsu-no koppu\/} `1-piece of cup' \\ \into\
{\it 1 cup\/} ({\sc general}) \label{class2}
\end{exe}

However if N is a noun that denotes an attribute, such as {\sc price}
or {\sc weight}, then the translation process becomes more
complicated.  In the simplest case the noun phrase {\it `XC-no-N'\/}
should be translated as though the classifier were a normal noun,
giving `{\it the\/} N {\it of\/} X C', for example (\ref{class3}), (\ref{class4}).

\begin{exe}
\ex     {\it 1-pai-no nedan\/} `1-cup of price' \\ 
\into\  {\it the price of 1 cup} \label{class3}
\ex {\it 1-tsu-no nedan [-ha 10en da]\/} `1-piece of price [-{\sc top} 
10 yen is]' \\ \into\  {\it the price of 1 (thing) [is 10 yen]}
\label{class4}
\end{exe}

In other words, if N has the attribute {\sc amount} then the noun
phrase should normally be translated as though C were not a
classifier.  The interpretation of C is, however, ambiguous.  C could
be used as a classifier with the amount N in its scope
(\ref{s:class5}), or C could have anaphoric reference
(\ref{s:class6}).  {\bf ALT-J/E} chooses the interpretation shown in
example (\ref{s:class6}) as its default.

\begin{exe}
\ex     {\it 1-sh\=u-no nedan\/} `1 kind of price' \\ \into\ 
 {\it 1 kind of price} \label{s:class5}
\ex     {\it 1-sh\=u-no nedan\/} `1 kind of price' \\ \into\  
{\it the price of 1 kind [of something]} \label{s:class6}
\end{exe}

Further, when N is an attribute and C measures the same attribute, the
interpretation is again different.  For example, if C measures N's
attribute then the resulting noun phrase will be indefinite by
default: {\it a height of 10m\/} or {\it a price of 10 yen}.  However
if the noun phrase is used ascriptively then it should be converted
either to an adjective {\it it is 10m high\/} or a prepositional
phrase {\it it is 10 yen in price}.  Finally, if a noun phrase of this
type is used to modify another noun then it needs to be converted to
an adjective {\it a 10m high building\/} or a post modifying
prepositional phrase {\it a chocolate 10 yen in price}.  

\medskip

The combinations of nouns and classifiers mentioned above can all be
translated by the machine translation system {\bf ALT-J/E} using the
analysis of classifiers presented in this paper in combination with a
semantic hierarchy of 2,800 categories common to all nouns, as
described in \newcite{Ikehara:1991}.  The particle {\it no\/} `of',
has many possible interpretations, \newcite{Shimazu:1987} identify
five main types of {\it A-no-B\/} expressions, and some 80 sub
types. Our analysis cuts across Shimazu et al.'s types, including at
least three of the subtypes, and also makes clear some relations that
are not explicitly named.

\section{Comparisons with other Analyses}
\label{sec:other}

We summarize our analysis of classifiers in Table~\ref{tab:overall}.
Our analysis was based mainly on the properties of the generated
English, so is naturally quite close to the division of partitive
nouns proposed by \newcite{Quirk:1985}.  The analysis is also quite
close to those proposed by \newcite{Kamei:1995} for Japanese and
Sornlertlamvanich et al. (1994) for Thai.  This supports Allan's
\shortcite{Allan:1977} assertion that ``diverse language communities
categorize perceived phenomena in similar ways''.  The different
analyses are compared in Table~\ref{tab:comp}.

We make the distinction between classifiers of frequency and other
{\sc unit} classifiers by using our general semantic hierarchy.
Sornlertlamvanich et al.'s {\sc verbal} classifiers ``any classifier
which is derived from a verb [\ldots] /kraad haa muan/ `five rolls of
paper'.'' can be included in the {\sc metric} category, although it
may be the case that they have a different part of speech in Thai.
\newcite{Kamei:1995} put {\sc unit} classifiers into two classes:
`Counting Total Amount': {\it 3kg of sugar\/} and `Counting an
Attribute Value': {\it a speed of 60mph}.  This distinction belongs to
the interpretation of the classifier in context, rather than its
inherent properties, so we feel the distinction should be made during
processing, as described in Section~\ref{sec:when}, rather than as
part of the analysis of the classifiers themselves.

\section{Conclusion}
\label{sec:conc}

In this paper we present an analysis of classifiers, suitable for use
in a Japanese-to-English machine translation system.  We divide
classifiers into four major types: {\sc unit}, {\sc metric}, {\sc
  group} and {\sc species}.  {\sc unit} classifiers are further
divided into {\sc general}, {\sc typical} and {\sc special}, while
{\sc metric} classifiers are divided into {\sc measure} and {\sc
  container} classifiers.  The analysis is based on characteristics
peculiar to Japanese and English, as well as the differences between
them.  The resulting analysis is shown to be similar to one proposed
for Thai, an unrelated language, suggesting that it may be more widely
applicable.

The analysis has been implemented in NTT's Japanese-to-English machine
translation system {\bf ALT-J/E} since 1994.  It makes possible a
uniform and straightforward treatment of noun phrases headed by
classifiers.

Further work remains to be done in examining the distribution of
classifiers in different domains, and possibly identifying classifiers
automatically.

\bibliographystyle{fullname}

\end{document}